\documentclass[conference]{IEEEtran}
\usepackage{amsmath}
\usepackage{amsfonts}
\usepackage{algorithm}
\usepackage[noend]{algpseudocode}
\usepackage{algorithm}
\usepackage{booktabs}
\usepackage{array}
\usepackage[numbers]{natbib}
\usepackage{multirow}
\usepackage[labelformat=simple]{subcaption}
\usepackage{graphicx}
\usepackage{listings}
\usepackage[utf8]{inputenc}
\usepackage[T1]{fontenc}

\begin{document}
\bstctlcite{IEEEexample:BSTcontrol}

\title{Enhancing the Monte Carlo Tree Search Algorithm\\for Video Game Testing
}

\author{\IEEEauthorblockN{Sinan Ariyurek}
\IEEEauthorblockA{\textit{Graduate School of Informatics} \\
\textit{Middle East Technical University} \\
Ankara, Turkey \\
sinan.ariyurek@metu.edu.tr}
\and
\IEEEauthorblockN{Aysu Betin-Can}
\IEEEauthorblockA{\textit{Graduate School of Informatics} \\
\textit{Middle East Technical University} \\
Ankara, Turkey \\
betincan@metu.edu.tr}
\and
\IEEEauthorblockN{Elif Surer}
\IEEEauthorblockA{\textit{Graduate School of Informatics} \\
\textit{Middle East Technical University} \\
Ankara, Turkey \\
elifs@metu.edu.tr}
}

\maketitle

\begin{abstract}
In this paper, we study the effects of several Monte Carlo Tree Search (MCTS) modifications for video game testing. Although MCTS modifications are highly studied in game playing, their impacts on finding bugs are blank. We focused on bug finding in our previous study where we introduced synthetic and human-like test goals and we used these test goals in Sarsa and MCTS agents to find bugs. In this study, we extend the MCTS agent with several modifications for game testing purposes. Furthermore, we present a novel tree reuse strategy. We experiment with these modifications by testing them on three testbed games, four levels each, that contain 45 bugs in total. We use the General Video Game Artificial Intelligence (GVG-AI) framework to create the testbed games and collect 427 human tester trajectories using the GVG-AI framework. We analyze the proposed modifications in three parts: we evaluate their effects on bug finding performances of agents, we measure their success under two different computational budgets, and we assess their effects on human-likeness of the human-like agent. Our results show that MCTS modifications improve the bug finding performance of the agents.
\end{abstract}

\section{Introduction}

The success of a video game can be attributed to various qualities, but bugs. The bugs found after release decreases the overall user experience and increases the budget spent on the game. To decrease the number of bugs, video game companies employ tremendous test efforts. However, video game testing is challenging as the game requirements change frequently \cite{Santos:2018}. The change in the game requirements requires repeating the tests and conducting new tests. Consequently, researchers proposed techniques to automate the video game testing such as scenario-based testing \cite{Cho:2010}, regression testing using record and replay segments \cite{Ostrowski:2013}, generating test sequences from UML and state diagrams \cite{Iftikhar:2015}, creating a Petri net of the game and producing sequences to be tested \cite{Becares:2016}, and employing reinforcement learning (RL) to expeditiously test an adventure game \cite{Pfau:2017}. Nonetheless, these approaches do not either provide an overall game testing experiment, or an automated oracle, or an intelligent tester agent, or comparison with human testers.

In our previous work \cite{Ariyurek:2019}, we generated test goals for Sarsa and MCTS agents to play the game with the purpose of testing the game (see Section \ref{subsec:previous_work}). In our experiments, we used Sarsa($\lambda$) \cite{Sutton:2018}, and MCTS with transpositions and knowledge-based evaluations (KBE) \cite{Browne:2012}. We used the GVG-AI framework to create testbed games that contain bugs. We conducted the experiments using these games, and our agents achieved comparable bug finding percentages with the human testers. Additionally, our experiments revealed that the stochasticity of MCTS is beneficial in bug finding. Therefore, in this paper, we investigate MCTS modifications and examine the consequences of different computational budgets for game testing purposes.

MCTS modifications are used by several researchers. In GVG-AI, several enhancements \cite{Perez:2014a}, \cite{Frydenberg:2015}, \cite{Soemers:2016}, \cite{Ilhan:2017} are employed to increase the performance of the Vanilla MCTS. Moreover, the authors \cite{Frydenberg:2015}, \cite{Soemers:2016} noted that not every enhancement has equal contribution to the performance. Furthermore, in board games, different MCTS enhancements \cite{Powley:2014} were favored in different games. Therefore, in this study, we experiment with several MCTS modifications and we compare them under two distinct computational budgets. Our aim is to analyze their impact on bug finding performances of our agents. In this regard, we propose to use 6 different enhancements, and within these enhancements, we introduce a new tree reuse strategy.

This paper is structured as follows: Section \ref{sec:preliminaries} gives preliminary information about MCTS, GVG-AI, and our previous work. Section \ref{sec:modifications} presents the considered modifications and their use in related research. The details of our experiments are given in Section \ref{sec:experiments}, and Section \ref{sec:results} presents the results. Section \ref{sec:discussion} discusses the outcome of the strategies used and their contributions. Section \ref{sec:conclusion} presents the conclusion and proposes further enhancements for future work.

\section{Preliminaries} \label{sec:preliminaries}

The following subsections introduce the preliminary material, as follows: MCTS, GVG-AI, and our previous work.

\subsection{Monte Carlo Tree Search}

Monte Carlo Tree Search (MCTS) \cite{Browne:2012} is a search algorithm that builds a tree to get the best available action. MCTS consists of four consecutive steps: selection, expansion, simulation, and backpropagation. These steps are executed iteratively until a certain condition is met. This condition can be a computational budget or finding the desiderata. The selection step chooses a node based on a Tree Policy. Eq. \ref{eq:ucb1} shows Upper Confidence Bounds (UCB1) which is a well-known approach. $\tilde{X_i}$ is the average score of the $i^{th}$ child, $C_p$ is the exploration constant, $n$ represents the visitation amount of the root, and $n_i$ is the visitation count of the $i^{th}$ child.

\begin{equation} \label{eq:ucb1}
    UCB1 = \tilde{X_i} + 2C_p \sqrt{\frac{2 \ln n}{n_i}}
\end{equation}

The expansion phase expands the search tree by adding one of the unexplored children of the selected node to the search tree. Simulation, starting from this unexplored child, generates actions to be taken based on a default policy. The score obtained from the state reached at the end of the simulation is backpropagated from the unexplored child up to the root. These four steps are executed in succession until the computational budget expires. Afterward, a child of the root node which corresponds to the best available action is returned.

\subsection{GVG-AI}

GVG-AI \cite{GVGAI:2016} is a framework that contains several two-dimensional games. There are more than 120 single-player games, including well-known games such as Mario, Sokoban, and Zelda. The game rules are written using a language called Video Game Description Language (VGDL) \cite{VGDL:2014}. The diversity of the games creates a challenging environment for general video game AI.

\subsection{Synthetic and Human-like Test Goals} \label{subsec:previous_work}

Game testing behavior is different from game playing, as a tester's aim is finding defects and interacting with the game to break it. A tester may have different goals besides finishing the game. In our previous research, we encapsulated these goals as test goals, and we proposed two different approaches to generate them, synthetic and human-like \cite{Ariyurek:2019}. 

A test goal $h$ consists of features $\phi$ and a criterion $c$ for each feature. $h{=}\{\phi_0,...,\phi_n,c_0,..c_n\}$ where $c_i$ is a positive rational number. Each feature stores a weight $w$ to define the reward obtained. Criterion defines a percentage for a feature such as the percentage of a wall to be tested and the percentage of space to be explored. If the agent interacts with a feature more than its criterion, the agent should receive less reward as the criterion is fulfilled. We implemented this behavior by using a dampening factor, which is used to diminish the reward obtained.

Moreover, we challenged our agents to test multiple goals in a game, but also preserve the goal order. Hence, the agent is given a sequence of goals $\mathcal{H} {=} (h_0,...,h_n)$, and starting from the first goal, the agent iterates them. When the agent accomplishes a goal, the agent passes on to the next goal.

The accomplishment of a goal is determined by a criteria threshold $c_T$. The sequence generated by the agent is checked to evaluate how much of the criteria are fulfilled. This evaluated score is then compared with the criteria threshold to determine whether the agent has reached the goal.

A game can be represented with a graph, where nodes are the states of a game, and edges are the actions that progress the story. We generated paths using this graph and a graph coverage criterion. Since playing only these paths corresponds to testing the valid game paths, we modified these paths to examine the effects of unintended game transitions. For a game where the player has to pick up the key to go through the locked door, two examples for unintended game transitions are attacking the key or trying to move through the locked door without picking the key. We created synthetic goals using these game paths and modified game paths. We also created baseline goals by only using the game paths.

We introduced multiple greedy-policy inverse reinforcement learning (MGP-IRL) to extract test goals from the collected human tester trajectories. We called the test goals obtained by this approach as human-like test goals.

Lastly, we introduced the test state, which is a supplementary state to the game state that holds executed interactions. In grid games, interactions occur between sprites. When the avatar attacks a wall, for example, we formulate this interaction and store it in the test state.

In this paper, an agent that uses a synthetic test goal is a synthetic agent, an agent that uses a 2baseline test goal is a baseline agent, and an agent that uses a human-like test goal is a human-like agent.

\section{MCTS Modifications} \label{sec:modifications}

Researchers designed various enhancements to increase the performance of a game-playing agent. However, we are interested in enhancing the bug-finding performance of our MCTS agent. In this section, we present the modifications we use in our tester agent. These modifications are presented by reviewing their uses in game-playing research. These modifications are namely: Transpositions, Knowledge-Based Evaluations, Tree Reuse, MixMax, Boltzmann Rollout, Single Player MCTS, and Computational Budget \cite{Browne:2012}.

\subsection{Transpositions}

In MCTS, the space of the game is explored as a tree, which can lead to having multiple nodes for a game state. Childs et al. \cite{Childs:2008} introduced transpositions in MCTS. Transposition tables (TT) promote sharing information between nodes of a tree that correspond to the same game state. The authors used this shared information to calculate the UCB1 value of a node, and they proposed three methods for this calculation. Perez et al. \cite{Perez:2014b} used TT for the Deep Sea Treasure game, Świechowski et al. \cite{Swiechowski:2016} used TT in General Game Playing (GGP), and Choe and Kim \cite{Choe:2019} used TT for Hearthstone. Xiao et al. \cite{Xiao:2018} used the feature representation of a state to query similar states in memory.

We use transposition tables since it is an effective method. In our implementation, an entry in the TT stores the information corresponding to a node in the search tree. The tree node only holds a pointer to an entry in the table. In the selection step, the information stored in the TT is used. In the backpropagation step, the information corresponding to the nodes, starting from the simulated node to the root node is updated. During the selection phase, the values stored in the table are used. The TT is not used during rollouts.

\subsection{Knowledge-Based Evaluations}

In GVG-AI, it is often difficult to find a terminal state or even a state that changes the game score, which may cause the MCTS agent to behave randomly. Powley et al. \cite{Powley:2014} introduced enhancements to exploit the episodic nature of games. Their information capture and reuse technique are found beneficial in games such as Dou Di Zhu, Hearts, and Othello. Soemers et al. applied 8 modifications to the open-loop implementation of MCTS. These modifications, one of which is KBE, are tested in the GVG-AI corpus, and the authors state that KBE modification is prominent. İlhan and Etaner-Uyar \cite{Ilhan:2017} combined MCTS with temporal difference learning in GVG-AI games to exploit domain knowledge using past experience. Silver et al. \cite{Silver:2016} trained a value network in AlphaGo to effectively evaluate the state of Go, and AlphaGo using this value network in MCTS beat the world champion.

In game testing research, the game being tested may contain bugs, and these bugs may prevent the agent from reaching a terminal state. Furthermore, terminal states or the points received from the game can be deceptive for the game testing agents, as losing the game is also an objective. Hence, we use the KBE to direct an agent.

\begin{equation} \label{eq:kbe}
    \textsc{Eval}_{\textsc{KBE}}(s, a) = \phi_{(s,a)} \cdot (w d) + (w_{h})^{f_{(s,a)}} - (w_{h})^{f_{s}}
\end{equation}

We evaluate the state using the Eq. \ref{eq:kbe}. $\phi_{(s,a)}$ with the features that are seen in state $s'$ resulting from taking an action $a$ on state $s$, and $w$ is the weights of these features. $w_h$ is the weight of completing a goal and $f_{s}, f_{(s,a)} \in [0, 1]$ represents the fulfillment amount of the goal criterion in state $s$ and $s'$, respectively. Note that, $f_{s}{=}0$ if $f_{s}$ is less than $c_T$. $d$ represents the dampening factor for the weights of the features that surpass their criteria. We employ this enhancement in all of our MCTS agents.

The list of the parameters and their values are as follows: Criteria threshold $c_T{=}0.01$, goal reward $w_{h}{=}10$, and the features observed but do not exist in the features $w{=}-1$.

\subsection{Tree Reuse}

Tree reuse strategy uses the previously generated tree to guide the forthcoming MCTS runs. Moreover, pruning this tree is as simple as selecting the subtree of the selected child. Santos et al. \cite{Santos:2017} employed tree reuse for MCTS agent in Hearthstone. Pepels et al. \cite{Pepels:2014} proposed a decaying tree reuse strategy in Ms. Pac-Man. Soemers et al. \cite{Soemers:2016} used this decaying tree reuse strategy in GVG-AI games, and this strategy is employed in Hearthstone \cite{Santos:2017}. In Ms. Pac-Man \cite{Pepels:2014} and GVG-AI \cite{Soemers:2016}, a decaying reuse strategy is employed. 

With transposition tables, e.g. the UCT3 \cite{Childs:2008} updates every node that precedes the simulated node, which is cumbersome and time-consuming. When the tree is reused, the complexity of updating a node increases as the game progresses. Furthermore, our proposed test state increases the number of states of a game. Consequently, reusing the whole search tree is not applicable, and we need to prune this tree. Pepels et al. \cite{Pepels:2014} used a rule-based method to remove the old nodes, and Powley et al. \cite{Powley:2017} proposed a node recycling method. Nevertheless, we propose a lightweight tree reuse method which presents effortless integration with transpositions, called as \textit{fast expansion}.

The \textit{fast expansion} uses the previously acquired tree in the selection and expansion phases. If the selected node exists in the previous tree, it is flattened and added to the current tree. The flattening process calculates the average score of $\tilde{X}$, and sets the visitation count of the node as $1$. The selection phase continues until it finds a node that does not exist in the previous tree. At this point, MCTS continues with simulation and backpropagation. This algorithm prunes the children that are not chosen in the selection and expansion steps. \textit{Fast expansion} supports acquiring the previous relevant knowledge and prevents bloating of previous visits and scores by flattening the nodes.

This approach can be perceived as remembering by doing. If MCTS repeats an action, its value is passed on to the next generation; otherwise, it is forgotten. Furthermore, this strategy can also be applied to graphs.

\subsection{MixMax}

Jacobsen et al. \cite{Jacobsen:2014} introduced MixMax to avert cowardly behavior in Super Mario Bros., and Khalifa et al. \cite{Khalifa:2016} used MixMax to enhance the human-likeness of an MCTS agent. Frydenberg et al. \cite{Frydenberg:2015} employed MixMax in GVG-AI games. The authors found mixed results for MixMax. We assume that a tester agent should be able to act more boldly as we want the agents to consider the paths that lead to a goal even though it is risky. Mixmax is used to blend the average score of $\tilde{X_i}$ in Eq. \ref{eq:ucb1} with the $max{X_i}$, shown in Eq. \ref{eq:mixmax}, where $Q$ is the mixing parameter.

\begin{equation} \label{eq:mixmax}
    max{X_i} \times Q + \tilde{X_i} \times (1 - Q)
\end{equation}

The win condition of our tester agent is the accomplishment of a test goal. Since MixMax modification supports choosing the risky move, it extends the possibility of pursuing this path. In the experiments, $Q$ is chosen as $0.25$.

\subsection{Boltzmann Rollout}

In Vanilla MCTS, the simulation policy selects random actions in the rollouts. In GGP, Finnsson and Bjornsson used Gibbs sampling to calculate the probability of actions. The authors biased the simulation policy by selecting an action using the probabilities. Tak et al. \cite{Tak:2012} argued that this selection mechanism does not fix the selection probability of the best action. Therefore, they used $\epsilon$-greedy to fix this probability. Powley et al. \cite{Powley:2014} stated that $\epsilon$-greedy approach is better than Gibbs sampling in GGP. In Go, Silver et al. \cite{Silver:2009} used softmax to parameterize the simulation policy. In GVG-AI, Perez et al. \cite{Perez:2014a} used the learned experience to bias the rollouts. In this study, we use the Boltzmann rollout. The Boltzmann rollout is based on Boltzmann exploration strategy in RL \cite{Sutton:2018}.

\begin{equation} \label{eq:boltzmann}
    p(i) = \frac{e^{\beta v(s,a)}}{\sum\limits_{a'}{e^{\beta v{(s,a')}}}}
\end{equation}

$p(i)$ represents the probability of choosing the1 $i^{th}$ node. The Boltzmann beta $\beta$ in Eq. \ref{eq:boltzmann} controls the randomness of the move where $\beta = 0$ is the same as random rollout. This equation determines the probability of choosing the node in the rollout. The value $v$ is the score obtained from taking the action $a$, in our case $v(s,a)$, is calculated using $\textsc{Eval}_{\textsc{KBE}}(s, a)$ Eq. \ref{eq:kbe}. 

On the other hand, James et al. \cite{James:2017} investigated why better-informed rollouts often result in worse-performing agents. In their work, they described that heavy knowledge-based rollouts cause high-bias and low variance which are choices that can result in poor performance. Hence, in the experiments, we choose $\beta = 0.5$ to increase the randomness of the simulation policy.

\subsection{SP-MCTS}

Schadd et al. \cite{Schadd:2008} introduced Single-Player MCTS (SP-MCTS). SP-MCTS modifies the UCB1 term which represents the possible deviation of a node. This term offers finer control to the exploration/exploitation dilemma in which the nodes have varying results. The authors showed that their modification outperformed other methods such as IDA$^{*}$, which is a variant of A$^{*}$ that limits the depth of search and iteratively increases this depth until the criterion is achieved, in puzzle games.

\begin{equation} \label{eq:sp_ucb}
    \tilde{X_i} + C_p \sqrt{\frac{2\ln n}{n_i}} + \sqrt{\frac{\sum{x^2 - n_i \tilde{X_i} + D}}{n_i}}
\end{equation}

SP-MCTS adds a third term for finer control in exploitation/exploration dilemma (see the third term in Eq. \ref{eq:sp_ucb}). This third term tilts UCB1 in favor of the nodes that have a high variance, and $D$ is a large constant value to mark rarely explored nodes as uncertain. In the experiments, $D$ is chosen as $10000$.

\subsection{Computational Budget}

Although the computational budget is not a modification, it is a parameter of MCTS. Nelson \cite{Nelson:2016} examined various computational budgets for MCTS. The author found that in the GVG-AI framework after a certain computational budget, the win rate becomes stable. Baier and Winands \cite{Baier:2016} compared different time management strategies of MCTS in detail for five different board games. The experimental results of research on the computational budget are promising. Therefore, we would like to investigate the effect of the computational budget on bug finding behavior.

\section{Experiments} \label{sec:experiments}

We created three games, each consisting of four levels, using the GVG-AI framework. We inserted a total of 45 bugs to these games which are mostly accomplished by changing the VGDL code. The first game has a 6$\times$7 grid and it is called Game A. In this game, the player has to pick up the key and go through the locked door to finish the game. The second game has an 8$\times$9 grid and is called Game B. In this game, the player has to put down the fire by pushing a water bucket, and pick up the key to go through the locked door. The last game, Game C, has a 10$\times$11 grid. In this game, the key is broken into two pieces, and the player has to combine them by pushing them into each other, then pick up the key to go through the locked door. For these three games, we used a similar sprite set, but a different layout for each level of a game.

We collected a total of 427 trajectories from 15 different human participants who have various gaming and testing experience. The testers warmed up by playing example levels to get used to the game controls and the environment. During testing, the players were able to test the games in any order and any number of times. The tester trajectories are collected using the GVG-AI framework.

Our human-like test goals are generated using these collected trajectories. During tests, human-like agents used the human-like test goals, which are extracted on the other three levels of the same game. We generated synthetic test goals by sampling paths from the game graph of a level. This game graph is provided by the game developer. The paths are modified using the sprite set of this level. The unmodified test goals are used as baseline test goals. During the tests, the synthetic agent used the synthetic test goals, and the baseline agent used baseline test goals, which are specifically generated for that level.

We created five different MCTS agents using the modifications described in Section \ref{sec:modifications}, and for each level, we ran them five times. These agents are KBE-MCTS, FE-MCTS, MM-MCTS, BR-MCTS, and SP-MCTS (see Table \ref{table:agents} for the modifications). All of the MCTS agents used $\gamma{=}0.95$, and rollout depth of 6. Exploration term is set as $C_p{=}0.95$ in all MCTS agents except SP-MCTS, which is $C_p{=}3.0$. For the computational budget, we experimented with 40 and 300 milliseconds on i7-8750H (4.1 GHz) using a single core. After these agents generated the test sequences, each sequence is checked by an automated test oracle.

In this study, we asked the following research questions (RQ). RQ1: What is the impact of different computational budgets? RQ2: Which modifications enhance MCTS's bug finding performance? RQ3: What is the bug finding performances compared to Sarsa($\lambda$) that uses the same test goals? RQ4: What is the effect of modifications on human-like behavior? The answers to these research questions and the results of the experiments are presented in the next section.

\renewcommand{\arraystretch}{0.7}

\begin{table}[]
\centering
\captionsetup{justification=centering}
\caption{Modifications Used in MCTS Agents}
\begin{tabular}{llccccc}
    &  & \multicolumn{5}{c}{ \textbf{Agents}}                                                                                                                                                                                                                                                                                                                                    \\ 
\cline{3-7} \\[0.02cm]
    &  & \multirow{3}{*}{\begin{tabular}[c]{@{}c@{}}KBE\\-\\MCTS \end{tabular}} & \multirow{3}{*}{\begin{tabular}[c]{@{}c@{}}FE\\-\\MCTS \end{tabular}} & \multirow{3}{*}{\begin{tabular}[c]{@{}c@{}}MM\\-\\MCTS \end{tabular}} & \multirow{3}{*}{\begin{tabular}[c]{@{}c@{}}BR\\-\\MCTS \end{tabular}} & \multirow{3}{*}{\begin{tabular}[c]{@{}c@{}}SP\\-\\MCTS \end{tabular}}  \\
    &  &                                                                        &                                                                       &                                                                       &                                                                       &                                                                        \\
    &  &                                                                        &                                                                       &                                                                       &                                                                       &                                                                        \\
\textbf{Modifications} &  & \multicolumn{1}{l}{}                                                   & \multicolumn{1}{l}{}                                                  & \multicolumn{1}{l}{}                                                  & \multicolumn{1}{l}{}                                                  & \multicolumn{1}{l}{}                                                   \\ 
\hline
\\
\vspace{0.15cm}
Transpositions         &  & \checkmark                                                              & \checkmark                                                             & \checkmark                                                             & \checkmark                                                             & \checkmark                                                              \\
\vspace{0.15cm}
KBE                    &  & \checkmark                                                              & \checkmark                                                             & \checkmark                                                             & \checkmark                                                             & \checkmark                                                              \\
\vspace{0.15cm}
Tree Reuse             &  &                                                                        & \checkmark                                                             &                                                                       &                                                                       &                                                                        \\
\vspace{0.15cm}
MixMax                 &  &                                                                        &                                                                       & \checkmark                                                             &                                                                       &                                                                        \\
\vspace{0.15cm}
Boltzmann Rollout      &  &                                                                        &                                                                       &                                                                       & \checkmark                                                             &                                                                        \\
SP-MCTS                &  &                                                                        &                                                                       &                                                                       &                                                                       & \checkmark                                                             
\end{tabular}
\label{table:agents}
\end{table}
\renewcommand{\arraystretch}{1.0}

\section{Results} \label{sec:results}

Table \ref{table:results} presents the results of our experiments. All of the values that are shown with intervals are in the confidence interval of $0.95$. For counting the number of bugs found, if there are multiple occurrences of the same bug, it is counted as one. As more than one testers tested a game, \textit{Combined} indicates all of the bugs found by these testers when their results are merged, and \textit{Individual} implies bugs found by each agent. As we aim to create tester agents, we also value the agents who find most bugs with shorter test sequences within the shortest computational budget. Lastly, we used cross-entropy to compare the interactions executed by a human tester's trajectory with that of the human-like agent. The lower the cross-entropy, the more similar are the interactions.

\subsection{Game A}

Game A has a 6$\times$7 grid size. Table \ref{table:results} shows that human testers, when combined, were able to find 90\% of the bugs, whereas individual performance is almost half of this score. Human-like MCTS agents with 40ms computational budget were able to find all of the bugs, except BR-MCTS. All MCTS agents generated a similar length sequence, except BR-MCTS. Cross-entropy scores of MM-MCTS and SP-MCTS are lower than the other MCTS agents. Increasing the computational budget increased the bug finding percentages and decreased the sequence lengths. Cross-entropy scores also decreased for every agent except FE-MCTS. The synthetic agent with a 40ms computational budget was not able to find all the bugs. SP-MCTS has the highest bug finding score and FE-MCTS has the lowest sequence length. The increase in the computational budget affects FE-MCTS and MM-MCTS positively. Baseline MCTS agents found at most 44\% of the bugs with 40ms computational budget, and an increase to the computational budget decreased the bug finding percentage to 40\%. Overall human-like MCTS scores are similar to human-like Sarsa($\lambda$), but synthetic Sarsa($\lambda$) score is better than synthetic MCTS.

\subsection{Game B}

\begin{figure}[]
    \centering
    \includegraphics[width=1.0\columnwidth]{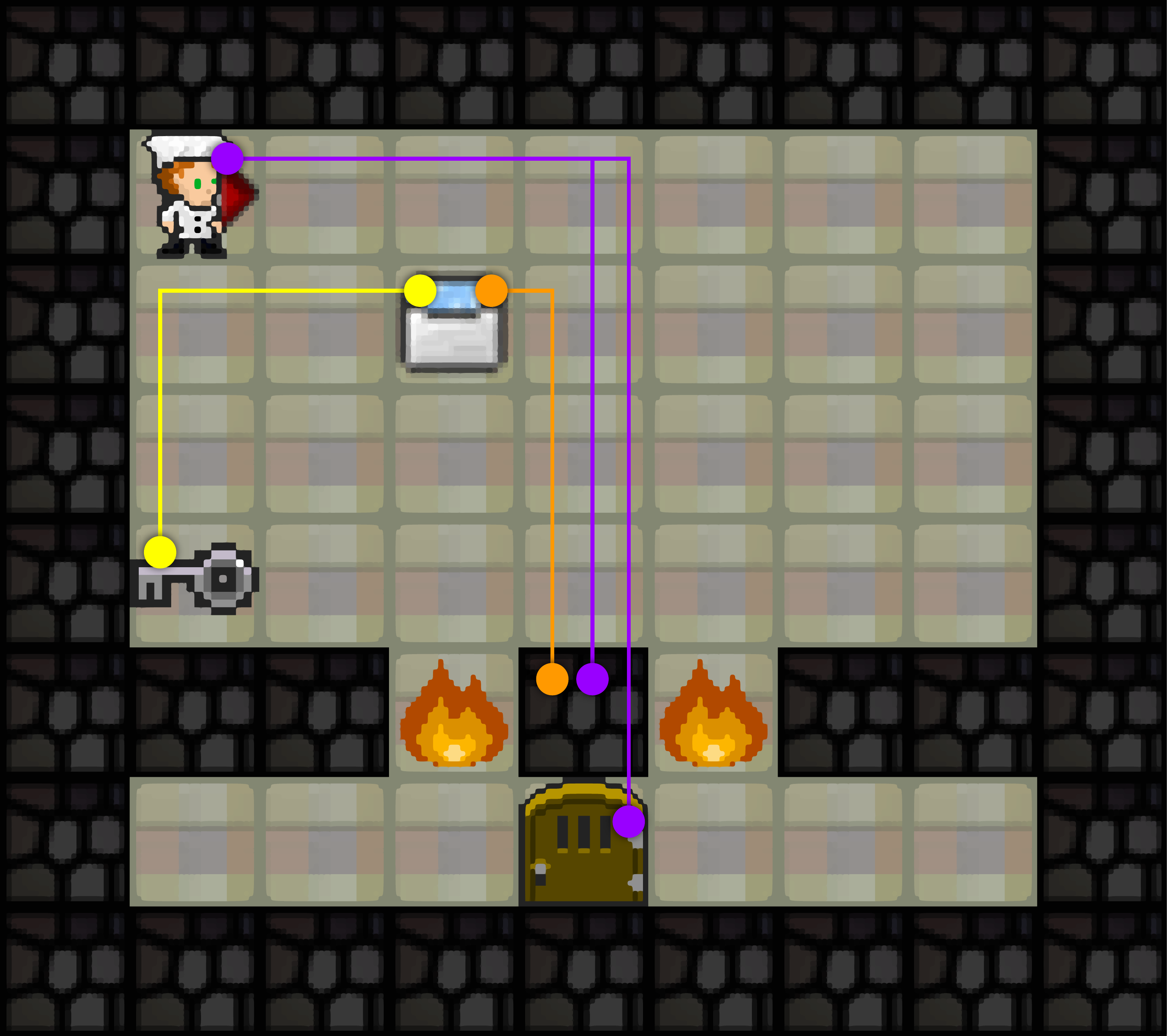}
    \caption{The paths that lead to four different bugs in Game B are shown with lines. Yellow Line: When the \textit{Avatar} pushes the \textit{Water Bucket} into \textit{Key}, the two sprites overlap. The rule to prevent this overlap is missing in VGDL, which is a discrepancy from the game design. Orange Line: The \textit{Avatar} can push the \textit{Water Bucket} into \textit{Wall}. The collision rule between this specific \textit{Wall} and \textit{Water Bucket} is missing in VGDL, which is another discrepancy from the game design. Purple Line: The \textit{Avatar} can move through the \textit{Wall} and can finish the game without picking up the \textit{Key}. These requirements exist in the game design, but they are not implemented in VGDL.}
    \label{fig:8x9_bugs}
\end{figure}

Game B has an 8$\times$9 grid size, as shown in Fig. \ref{fig:8x9_bugs}. Table \ref{table:results} shows that human testers, combined, were able to find all of the bugs. However, when they are evaluated individually, their scores are lower than Game A. In Game B, none of the agents were able to find all of the bugs. Under the 40ms computational budget, FE-MCTS found more bugs than other MCTS agents. The sequence length of all MCTS agents is similar except BR-MCTS. KBE-MCTS and then FE-MCTS has the lowest cross-entropies. The increase in computational budget decreased the cross-entropies of all agents. This increase also positively affected all agents except FE-MCTS. For synthetic agents, under both computational budgets, FE-MCTS found more bugs than other MCTS agents. Baseline scores of KBE-MCTS, FE-MCTS, and BR-MCTS are close and higher than MM-MCTS and SP-MCTS. The bug finding percentage of human-like Sarsa($\lambda$), and synthetic Sarsa($\lambda$) is higher than human-like MCTS, and synthetic MCTS, respectively.

\subsection{Game C}

Game C has the biggest grid size of all three games, which is 10$\times$11. Table \ref{table:results} shows that several MCTS agents were able to surpass the Sarsa($\lambda$). The individual bug finding performances of human testers are the lowest, and their combined performance is 90\%. Although synthetic MCTS agents do not surpass synthetic Sarsa($\lambda$) agents, the human-like MCTS agents surpass human-like Sarsa($\lambda$) agents and some, even surpass human testers. The increase in the computational budget also increases the performance of every MCTS agent except baseline BR-MCTS. Using 300ms computational budget, human-like FE-MCTS and SP-MCTS beat every other agent. The shortest trajectory amongst human-like agents is executed by MM-MCTS, and FE-MCTS has the lowest cross-entropy. Synthetic BR-MCTS has the best bug finding percentage amongst MCTS agents. The baseline agent was not able to find any bugs in some runs and was most efficiently played by FEBR-MCTS.

\renewcommand{\arraystretch}{1.25}
\setlength{\tabcolsep}{3.2pt}

\begin{table*}[]
% \small
\centering
\captionsetup{justification=centering}
\caption{Bug Finding Percentage, Trajectory Length, Cross-Entropy Results of Human Testers and Agents using 
Sarsa($\lambda$), KBE-MCTS, MM-MCTS, FE-MCTS, BR-MCTS, SP-MCTS obtained from Game A (6x7), Game B (8x9), and Game C (10x11).
The values shown with range have values Confidence Interval of 0.95.}
\begin{tabular}{cccccccccccc}
    \hline
    &\multicolumn{3}{c}{\textbf{Bug Finding Percentage \%}} & & \multicolumn{3}{c}{\textbf{Trajectory Length}} & & \multicolumn{3}{c}{\textbf{Cross-Entropy}}  \\    \cline{2-4} \cline{6-8} \cline{10-12}
    \multicolumn{1}{l}{\textbf{Tester}} & \textbf{\begin{tabular}[c]{@{}c@{}}Game A\\ (6x7)\end{tabular}} & \textbf{\begin{tabular}[c]{@{}c@{}}Game B\\ (8x9)\end{tabular}} & \textbf{\begin{tabular}[c]{@{}c@{}}Game C\\ (10x11)\end{tabular}} & & \textbf{\begin{tabular}[c]{@{}c@{}}Game A\\ (6x7)\end{tabular}} & \textbf{\begin{tabular}[c]{@{}c@{}}Game B\\ (8x9)\end{tabular}} & \textbf{\begin{tabular}[c]{@{}c@{}}Game C\\ (10x11)\end{tabular}} & & \textbf{\begin{tabular}[c]{@{}c@{}}Game A\\ (6x7)\end{tabular}} & \textbf{\begin{tabular}[c]{@{}c@{}}Game B\\ (8x9)\end{tabular}} & \textbf{\begin{tabular}[c]{@{}c@{}}Game C\\ (10x11)\end{tabular}} \\ \hline
    \multicolumn{3}{l}{\textbf{Humans}}     &       &      &    \\
    Combined & $90.0$    & $100.0$  & $90.0$ && $41.0-58.8$ & $38.8-49.7$ & $74.5-99.9$ && \\
    Individual & $42.7-56.0$ & $29.5-46.3$ & $26.7-43.7$ \\ \hline
    \multicolumn{3}{l}{\textbf{Sarsa($\lambda$)}}     &       &      &    \\
    Synthetic & $100.0$ & $76.2$   & $70.0$ && $31.6-50.1$ & $75.8-89.1$ & $129.0-148.4$ && \\
    Human-Like & $100.0$ & $90.5$   & $70.0$ && $39.3-48.0$ & $47.7-53.2$ & $97.9-109.6$ && $0.27-0.37$ & $0.57-0.69$ & $0.69-0.82$\\ 
    Baseline & $30.0$  & $42.9$   & $10.0$ && $6.8-14.8$ & $13.1-21.8$ & $30.2-65.3$ && \\ \hline
    \multicolumn{4}{l}{\textbf{Computational Budget 40ms}} & & \multicolumn{3}{c}{\textbf{Sequence Length}} \\\hline
    \multicolumn{1}{l}{\textbf{KBE-MCTS}}\\
    Synthetic & $84.0-90.0$ & $64.0-70.0$ & $36.0-50.0$ && $84.4-109.3$ & $98.4-108.0$ & $211.6-232.5$ && \\
    Human-Like &$86.0-100.0$ & $67.0-73.0$ & $60.0-68.0$ && $70.2-78.6$ & $61.3-66.5$ & $90.1-99.5$ && $0.72-0.82$ & $1.12-1.20$ & $1.15-1.22$\\
    Baseline & $20.0-20.0$ & $34.0-41.2$ & $10.0-10.0$ && $14.7-38.9$ & $52.0-70.7$ & $72.9-129.1$ && \\ \hline
    \multicolumn{1}{l}{\textbf{MM-MCTS}}\\
    Synthetic & $82.0-90.0$ & $46.0-58.6$ & $42.0-50.0$ && $98.5-129.2$ & $100.6-111.4$ & $199.8-219.4$ && \\
    Human-Like & $92.0-100.0$ & $74.6-83.0$ & $58.0-88.0$ && $72.1-80.2$ & $61.9-67.8$ & $74.5-83.5$ && $0.57-0.64$ & $1.24-1.33$ & $1.20-1.27$\\
    Baseline & $26.0-40.0$ & $9.0-17.0$ & $0.0-0.0$ && $19.2-59.0$ & $67.3-104.6$ & $91.8-132.6$ && \\ \hline
    \multicolumn{1}{l}{\textbf{FE-MCTS}}\\
    Synthetic & $84.0-90.0$ & $59.8-68.0$ & $52.0-64.0$ && $62.7-84.8$ & $97.0-106.0$ & $195.9-215.0$ && \\
    Human-Like & $92.0-100.0$ & $76.6-87.0$ & $72.0-80.0$ && $70.8-78.8$ & $63.2-68.7$ & $80.3-89.5$ && $0.61-0.70$ & $1.16-1.25$ & $1.18-1.25$\\
    Baseline & $20.0-28.0$ & $33.0-39.4$ & $10.0-10.0$ && $13.2-32.8$ & $58.7-85.1$ & $48.0-102.0$ && \\ \hline
    \multicolumn{1}{l}{\textbf{BR-MCTS}}\\
    Synthetic & $82.0-90.0$ & $57.8-65.0$ & $50.0-66.0$ && $96.7-128.4$ & $112.8-125.8$ & $207.1-228.2$ && \\
    Human-Like & $76.0-90.0$ & $76.0-82.0$ & $56.0-70.0$ && $89.0-99.8$ & $103.0-112.6$ & $117.1-130.6$ && $1.10-1.22$ & $1.22-1.31$ & $1.15-1.22$\\
    Baseline & $20.0-20.0$ & $32.0-40.4$ & $13.2-26.4$ && $22.9-82.6$ & $53.5-82.3$ & $36.7-76.4$ && \\ \hline
    \multicolumn{1}{l}{\textbf{SP-MCTS}}\\
    Synthetic & $84.0-96.0$ & $46.8-60.2$ & $38.0-58.0$ && $99.3-127.9$ & $107.2-118.8$ & $197.2-217.9$ && \\
    Human-Like & $94.0-100.0$ & $72.0-83.2$ & $70.0-92.0$ && $74.4-82.8$ & $61.6-67.5$ & $75.9-85.1$ && $0.58-0.66$ & $1.21-1.30$ & $1.20-1.27$\\
    Baseline & $26.0-44.0$ & $15.0-19.0$ & $0.0-6.0$ && $15.5-35.2$ & $65.3-105.5$ & $61.3-130.1$ && \\ \hline
    \multicolumn{9}{l}{\textbf{Computational Budget 300ms}}\\\hline
    \multicolumn{1}{l}{\textbf{KBE-MCTS}}\\
    Synthetic & $84.0-90.0$ & $61.0-71.0$ & $46.0-60.0$ && $76.5-97.8$ & $103.4-112.9$ & $219.9-239.0$ && \\
    Human-Like & $100.0-100.0$ & $75.6-84.0$ & $68.0-90.0$ && $63.5-71.1$ & $67.9-73.5$ & $109.3-118.5$ && $0.65-0.75$ & $1.00-1.07$ & $1.05-1.12$ \\
    Baseline & $20.0-26.0$ & $30.0-36.0$ & $10.0-16.0$ && $10.0-15.8$ & $54.9-77.5$ & $82.2-132.0$ && \\\hline
    \multicolumn{1}{l}{\textbf{MM-MCTS}}\\
    Synthetic & $84.0-96.0$ & $49.6-64.0$ & $48.0-70.0$ && $86.2-109.1$ & $95.9-106.3$ & $206.2-226.9$ && \\
    Human-Like & $100.0-100.0$ & $79.4-87.0$ & $68.0-92.0$ && $66.0-73.4$ & $64.2-69.7$ & $77.7-87.2$ && $0.57-0.65$ & $1.11-1.20$ & $1.21-1.28$\\
    Baseline & $30.0-38.0$ & $15.0-22.2$ & $0.0-12.0$ && $18.7-55.1$ & $52.8-79.0$ & $64.5-132.3$ && \\ \hline
    \multicolumn{1}{l}{\textbf{FE-MCTS}}\\
    Synthetic & $90.0-96.0$ & $60.6-74.0$ & $50.0-66.0$ &&  $49.6-64.3$ & $88.8-97.2$ & $202.8-220.0$ && \\
    Human-Like & $94.0-100.0$ & $76.8-82.2$ & $76.0-94.0$ && $47.4-52.8$ & $53.4-57.8$ & $105.7-115.0$ && $0.75-0.85$ & $0.98-1.06$ & $0.98-1.05$\\
    Baseline & $24.0-40.0$ & $35.8-42.0$ & $10.0-10.0$ && $10.8-18.1$ & $38.4-55.9$ & $89.7-131.6$ && \\ \hline
    \multicolumn{1}{l}{\textbf{BR-MCTS}}\\
    Synthetic & $80.0-88.0$ & $66.0-72.0$ & $60.0-66.0$ && $61.7-85.3$ & $77.7-86.8$ & $198.8-218.4$ && \\
    Human-Like & $84.0-90.0$ & $76.8-82.2$ & $74.0-80.0$ && $77.0-85.6$ & $88.7-97.0$ & $116.6-128.4$ && $0.93-1.05$ & $1.00-1.08$ & $1.08-1.15$\\
    Baseline & $22.0-34.0$ & $21.4-32.2$ & $2.0-14.0$ && $15.1-31.4$ & $42.4-64.1$ & $54.4-117.4$ && \\ \hline
    \multicolumn{1}{l}{\textbf{SP-MCTS}}\\
    Synthetic & $84.0-96.0$ & $44.0-58.4$ & $46.0-62.0$ && $83.5-108.4$ & $105.1-115.7$ & $202.4-223.8$ && \\
    Human-Like & $100.0-100.0$ & $81.0-85.0$ & $76.0-94.0$ && $68.5-75.9$ & $63.9-69.3$ & $82.5-91.6$ && $0.53-0.61$ & $1.12-1.20$ & $1.19-1.26$\\
    Baseline & $30.0-38.0$ & $15.0-20.6$ & $0.0-6.0$ && $16.4-42.8$ & $48.9-76.1$ & $57.2-130.6$ && \\ \hline

\end{tabular}
\label{table:results}
\end{table*}

\section{Discussion} \label{sec:discussion}

In this paper, we experimented with several modifications to MCTS for creating a better tester agent. We experimented with these modifications on three games with 45 bugs, evaluated their effects on bug finding performance, how they are affected by the computational budget, and compared these findings with the human testers and an agent using Sarsa($\lambda$).

To address RQ1, we discuss the effect of increasing the computational budget on the MCTS modifications. For human-like agents, our experiments reveal that the increase in computational budget positively affected most of the MCTS agents. Amongst those, KBE-MCTS benefits the most. Therefore, we can state that KBE-MCTS was not able to explore the tree using a 40ms computational budget. SP-MCTS and MM-MCTS are also affected positively, but not as much as KBE-MCTS. However, with additional computation, they become more stable agents, as their confidence interval shrinks. BR-MCTS has a better performance than KBE-MCTS in Game B, and Game C using a 40ms computational budget. However, the increase in the computational budget reverses the situation. The bias in the rollouts of BR-MCTS limits its upper bound and becomes the most stable agent. The only advantage of FE-MCTS over KBE-MCTS is tree reuse. Tree reuse improves the bug-finding percentage considerably under 40ms, but with the increase in computational budget, KBE-MCTS outperforms FE-MCTS in Game A and Game B. Since these games are small compared to Game C, tree reuse starts to decrease the stochasticity. FE-MCTS also executes the shortest sequences under 300ms computational budget. For synthetic agents, the increase in computational budget increased the bug finding performance of all agents, which indicates that synthetic test goals are more difficult to reach compared to human-like test goals. The percentage of bugs that a baseline agent can find is limited to the bugs in the scenario, and baseline Sarsa($\lambda$) represents this percentage. In baseline FE-MCTS, the increase in the computational budget had a positive effect. For other baseline MCTS agents, the effect is perplexed. Furthermore, there are instances where a baseline MCTS agent surpasses the baseline Sarsa($\lambda$). This bug-finding performance boost is also due to the stochasticity of the MCTS, which also contributed to finding the fake walls in \cite{Machado:2018}.

We address RQ2 by comparing the effects of modifications on bug finding performances. For human-like agents, SP-MCTS with 300ms computational budget is the overall best. Although FE-MCTS can reach the same upper bound and even exceed that bound, FE-MCTS achieves this performance when we consider both computational budgets. However, for these instances, the sequence lengths of SP-MCTS are shorter than FE-MCTS. MM-MCTS has more variance than the other agents in Game C. This variance can be explained by the MixMax modification. The upper bounds of MM-MCTS and SP-MCTS are close, but since SP-MCTS explores more, it guarantees a higher lower bound. BR-MCTS is the least successful human-like agent, but it is stable. On the other hand, when we look at synthetic agents, SP-MCTS is one of the least successful agents, and BR-MCTS starts to excel. This indicates that synthetic test goals are located deeper in the tree, compared to human-like goals. Therefore, MixMax, Tree Reuse becomes useful. BR-MCTS is also useful as these goals can be found during biased rollouts. There is no clear winner in synthetic MCTS, but FE-MCTS is promising.

To address RQ3, we compare the MCTS variants with Sarsa($\lambda$) using Table \ref{table:results}. In Game A, human-like agents using MCTS variants reach the bug finding percentage obtained with Sarsa($\lambda$), and they can achieve this with a 40ms computational budget, except BR-MCTS. In Game B, the human-like agent using MM-MCTS, FE-MCTS, and SP-MCTS is 3-5\% behind of Sarsa($\lambda$). In Game C, the MM-MCTS, FE-MCTS, and SP-MCTS with 40ms computational budget and all human-like MCTS agents with 300ms surpass the bug finding performance of Sarsa($\lambda$). These bug finding percentages show that human-like MCTS agents can compete with Sarsa($\lambda$) in bug-finding with the advantage of using a less computational budget. For synthetic agents, we observe that bug finding performance of Sarsa($\lambda$) is better. Nevertheless, in Game A and Game B, FE-MCTS; in Game C, MM-MCTS are the best competitors. Furthermore, for every game, we can see that Sarsa($\lambda$) produces shorter sequences and these sequences are more human-like compared to MCTS agents. For the baseline agent, FE-MCTS with 300ms computational budget performs closest to Sarsa($\lambda$), thanks to the tree reuse modification.

We address the RQ4 by comparing the cross-entropy scores in Table \ref{table:results}. There is a direct relation with the human-likeness of KBE-MCTS, SP-MCTS, and MM-MCTS with the computation budget, but not for FE-MCTS, and MM-MCTS. However, we cannot state that if an agent performs closer to the original human it will find more bugs. The heuristics learned from human testers provide the goals to test the game, and any randomization added while generating a sequence will decrease the similarity. However, due to randomization, different runs can find distinct bugs.

\section{Conclusion} \label{sec:conclusion}

In this paper, we employed several modifications to the MCTS algorithm to evaluate their effects on finding bugs. In this regard, we proposed to use transpositions, knowledge-based evaluations, tree reuse, MixMax, Boltzmann rollouts, and SP-MCTS. We exercised these modifications in three games. Our synthetic and human-like test goals were exercised using MCTS to generate sequences that were later replayed in the game to check for bugs with our oracle. We employed two different computational budgets, 40 and 300 milliseconds, to better understand the effect of timing on these modifications.

Our results show that the modifications are useful, but the effectiveness of modification depends on the type of the agent. From our experiments, we found that for the synthetic agent MM-MCTS and FE-MCTS performed better, but BR-MCTS had a better lower bound score with a shorter sequence. FE-MCTS with 300ms computational budget was a better baseline agent than the other baseline MCTS. For human-like agents, SP-MCTS performed solid within both computational budgets, and FE-MCTS was a close contender.

In the future, we would like to experiment with reuse strategies for MM-MCTS and SP-MCTS. MM-MCTS and FE-MCTS are the best performing synthetic MCTS agents, so their combination may beat synthetic Sarsa($\lambda$). Integrating tree reuse to SP-MCTS may create a more powerful human-like agent. Furthermore, we would like to extend the experiments with various GVG-AI games.

\renewcommand{\bibfont}{\small}
\bibliographystyle{IEEEtran}
\bibliography{IEEEabrv,References}

\end{document}